# Anomalous State Sequence Modeling to Enhance Safety in Reinforcement Learning

**Leen Kweider[1], Maissa Abou Kassem[1], and Ubai Sandouk[2]**

[1]Department of Artificial Intelligence, Faculty of Information Technology Damascus University, Damascus, Syria
[2]Department of Software Engineering, Faculty of Information Technology Damascus University, Damascus, Syria

Corresponding author: Leen Kweider (e-mail: leenkweider@gmail.com).

**Abstract**

The deployment of artificial intelligence (AI) in decision-making applications requires ensuring an appropriate level of safety and reliability, particularly in changing environments that contain a large number of unknown observations. To address this challenge, we propose a novel safe reinforcement learning (RL) approach that utilizes an anomalous state sequence to enhance RL safety. Our proposed solution Safe Reinforcement Learning with Anomalous State Sequences (AnoSeqs) consists of two stages. First, we train an agent in a non-safety-critical offline 'source' environment to collect safe state sequences. Next, we use these safe sequences to build an anomaly detection model that can detect potentially unsafe state sequences in a 'target' safety-critical environment where failures can have high costs. The estimated risk from the anomaly detection model is utilized to train a risk-averse RL policy in the target environment; this involves adjusting the reward function to penalize the agent for visiting anomalous states deemed unsafe by our anomaly model. In experiments on multiple safety-critical benchmarking environments including self-driving cars, our solution approach successfully learns safer policies and proves that sequential anomaly detection can provide an effective supervisory signal for training safety-aware RL agents.

## INTRODUCTION

Ensuring the safety and reliability of Artificial Intelligence (AI) is critical, particularly in decision-making applications in dynamically changing real-world environments that introduce numerous unknown observations that may entail high costs or risks, making it crucial to ensure that AI systems can operate safely and efficiently. Reinforcement Learning (RL) is a popular technique used in decision-making applications. However, its deployment in safety-critical environments poses additional difficulties.

One of the primary challenges in deploying RL in real-world scenarios is the high training cost in real environments. Traditional RL approaches rely on trial and error, which can lead to undesirable outcomes in situations in which safety is paramount. Moreover, RL algorithms typically focus on maximizing rewards, which may not align with the safety considerations. Therefore, there is growing interest in developing RL techniques that prioritize safety alongside performance.

Another challenge of traditional RL algorithms is that they often assume that the environment operates as expected without considering the possibility of anomalous states or actions. However, in real-world scenarios, anomalies can arise due to various factors such as sensor failures, adversarial attacks, or changes in the environment.

To address these limitations, we propose a novel safe RL approach named Safe Reinforcement Learning with Anomalous State Sequences (AnoSeqs), which leverages anomalous state sequences as a measure of unsafety. By considering the sequence of states over time, our approach can identify potential safety risks that may not be apparent from single-state analysis. In particular, using state sequences rather than individual states for anomaly detection is essential for safety-critical applications. State sequences capture the temporal dynamics of the environment and can reveal patterns of unsafety or anomalous behavior that may not be apparent from single states. By considering the sequence of states over time, the proposed approach can better detect potential hazards and take appropriate actions to mitigate them.

Our proposed method consists of two stages. First, we train an agent in a non-safety-critical offline 'source' environment to collect safe state sequences. We then employed these safe sequences to build a sequential anomaly detection model that can detect potentially unsafe states in a safety-critical target environment. Finally, we use the estimated risk (anomaly score) from the anomaly detection model to train a risk-averse policy in the target environment, adjusting the reward function to penalize the agent for visiting states deemed unsafe by the anomaly detection output. We evaluate our approach using multiple safety-critical benchmarking environments, including self-driving cars. Our results demonstrate that incorporating sequential anomaly detection as an unsafety measure in RL leads to safer policies and more reliable decision-making processes. This study contributes to the development of practical and safe AI systems for real-world applications, ensuring that AI can effectively assist humans in various domains while minimizing potential risks.

Overall, this research contributes to the development of safer AI systems by introducing a practical and effective method for integrating sequence anomaly detection into RL algorithms.

## RELATED WORK

Several approaches have been proposed in recent years to enhance safety in reinforcement learning (RL) systems, particularly in dynamic environments with high uncertainty and potential risks. In this section, we present a summary of notable works and contributions to safe reinforcement learning. [5] provided a foundational perspective on safe reinforcement learning, this survey offers a classification of methods based on their approach to exploration and optimization and serves as a useful reference for understanding the early development of Safe Reinforcement Learning. More recently, [6] provided a broader view of safe RL. The authors attempted to create a bridge between control systems and RL. They categorized safety methods based on the kind of constraints they satisfy, including hard constraints, probabilistic constraints, and soft constraints.

The literature provides methods to improve RL safety at various stages of the RL process. We categorize these approaches based on where and how safety is integrated into the RL process, highlighting key contributions and methodologies. Related approaches include:

- **Optimization level safety:**

Traditional RL algorithms primarily focus on finding optimal control policies that guide actions to optimize a given criterion, typically cumulative rewards over time. However, these methods often neglect safety considerations, leading to potentially risky or unsafe actions. To address this issue, several approaches have been proposed to incorporate safety by modifying optimization criterion in the reinforcement learning algorithm.

One type of optimization criterion modification is the use of constrained optimization techniques, where the agent's policy is optimized subject to safety constraints defined based on domain-specific knowledge or predefined safety rules. Constrained Markov Decision Process (CMDP) was introduced by [7]. In this approach, the RL agent is trained to maximize its expected reward while respecting a set of safety constraints. These constraints ensure that the agent operates within acceptable boundaries, mitigating the risk of catastrophic outcomes.

Further studies have explored the use of safety constraints in RL, as explored by studies such as [8-11].

**Lagrangian algorithm [1]:** Lagrangian method is a mathematical technique used to solve optimization problems with equality and inequality constraints. In the context of RL, this method can be used to optimize the policy to maximize cumulative expected rewards while satisfying certain constraints. The Lagrangian approach allows for the incorporation of safety constraints dynamically during the learning process. By introducing a Lagrange multiplier, the constrained optimization problem can be rewritten to balance the trade-off between reward maximization and constraint satisfaction.

- **Incorporating Risk-Averse Policies:**

Another type of modifying optimization criterion is risk-sensitive criterion where risk-sensitive RL policies explicitly consider the uncertainty or risk associated with different actions. It aims to balance the trade-off between maximizing expected rewards and minimizing the potential negative consequences of risky actions. These policies employ various risk (or uncertainty) measures, these estimates are typically computed for the system dynamics or the overall cost function and leveraged to produce more conservative (and safer) policies, Risk measure can be defined as conditional value-at-risk (CVaR) or entropy regularization, or any other risk measures [12-14] to guide the agent's decision-making process towards less cost actions.

Researchers have also explored the use of learned risk measures or uncertainty risk measures that capture risk preferences directly from data or model uncertainty. These measures allow RL agents to adapt to varying risk preferences or handle situations where traditional risk measures may not be adequate. Risk can be defined as the probability of collision, uncertainty estimate, or probability of unsafe state-action pairs [15-19].

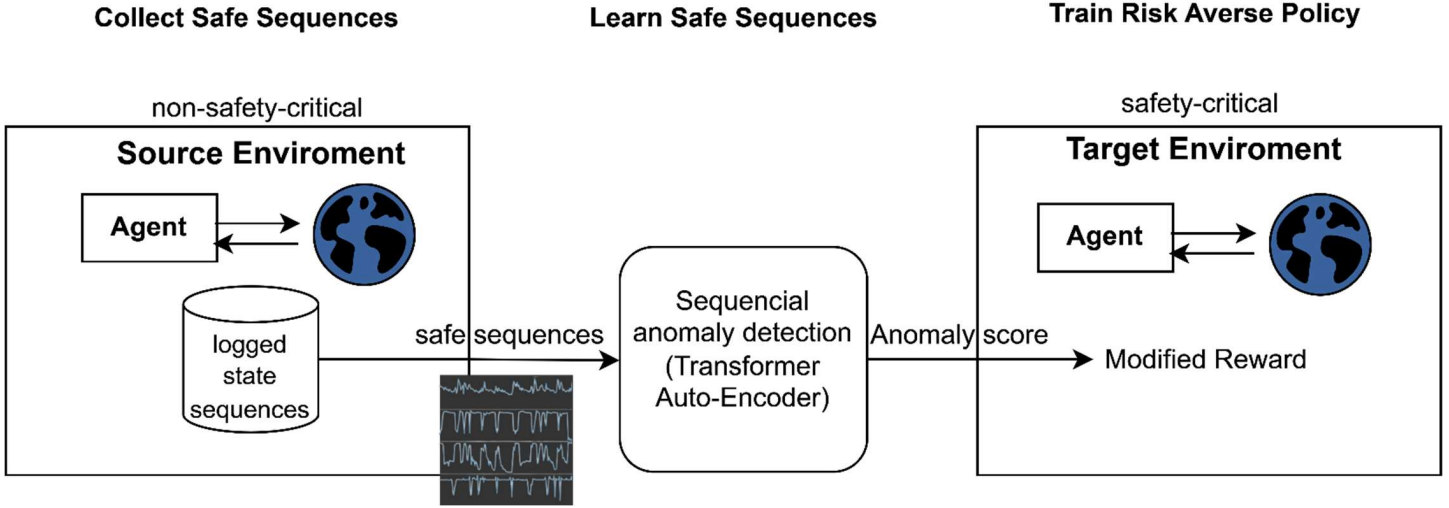

FIGURE 1 Proposed approach (AnoSeqs); In the first phase, an agent is trained in a simulated "source" environment without safety concerns to log all state sequences. Then get safe sequences data to train a transformer auto-encoder to learn the safe sequences. In the second phase, in a "target" environment where safety concerns matter; an agent is trained in a risk-averse policy where the state sequence anomaly is the estimated risk.

**Reward shaping algorithm [3, 4]:** Reward shaping algorithm is a technique used to accelerate the learning process in RL by providing additional rewards (or penalties) to guide the agent towards desirable behaviors. This is often done to make the learning process safer or to incorporate safety or other constraints directly into the reward function. In this context, the reward function is adjusted by adding a penalizing term for risk, which helps balance the trade-off between expected rewards and potential risks.

- **Safety Certificates and Filters:**

In fields like robotics and autonomous driving, specific safety measures are applied to policy outputs, such as safety certificates or filters that ensure the agent's actions remain within safe boundaries.

**Recovery RL Algorithm [2]:** Recovery RL algorithm involves a dedicated recovery policy to handle unsafe situations. In Recovery RL, a secondary recovery policy is developed alongside the primary task policy. The recovery policy is activated when the agent's actions are deemed unsafe according to a predefined safety criterion. This ensures that the agent can recover from unsafe states and continue its operation within safe limits. An example of this approach is demonstrated in. [2], where the recovery policy is used to navigate a two-dimensional maze without colliding with walls. The safety (risk) is evaluated for each action, and the recovery policy is executed if the risk exceeds a certain threshold.

- **Adaptation from Other Environments**

Domain adaptation and randomization techniques aim to transfer knowledge from a source domain to a target domain with different characteristics. In the context of safe RL, these techniques can be employed to adapt safety-related knowledge learned in one environment to a different real-world environment. Notable contributions in this area include the works by [18, 20-22].

- **Anomaly Detection in Reinforcement Learning**

Anomaly detection is a technique used in data analysis and machine learning that aims to identify patterns or instances that deviate significantly from the expected behavior within a dataset [23], The goal of anomaly detection is to distinguish between normal and abnormal data points, aiding in the identification of unusual events, errors, or any unexpected behavior that could potentially be risky.

Sequential anomaly detection is an extension of traditional anomaly detection that specifically focuses on detecting anomalies within sequential data, where the order and temporal dependencies of data points are important [24].

The integration of anomaly detection with reinforcement learning offers promising avenues for addressing safety concerns and ensuring robustness in RL applications. Recently, [25] in their paper "Towards Anomaly Detection in Reinforcement Learning" explored the integration of anomaly detection (AD) with reinforcement learning and investigated the implications of AD in the context of RL and how these two fields can mutually benefit from each other.

## PROPOSED METHODOLOGY

Our proposed approach Safe Reinforcement Learning with Anomalous State Sequences (AnoSeqs) utilizes sequential anomaly detection modeling in reinforcement learning as the measure of unsafety. As illustrated in Figure 1 the proposed approach (AnoSeqs) operates in two distinct phases: learn unsafe state sequence detection and train risk-averse policy.

- **Phase 1: Learn Unsafe State Sequence Detection**

In this phase, we first collect safe state sequences by training an RL agent to explore and navigate in an offline simulated non-safety-critical "source" environment and then learn the safe state sequences with an anomaly detection model.

**1- Collect Safe Sequences:**
We train a reinforcement learning agent, specifically an actor-critic reinforcement learning agent, in an offline simulated non-safety-critical "source" environment. The objective is to collect safe state sequences, which do not lead to unsafe states, agent damage, or episode termination. "Unsafe states" are defined as states where the agent violates predefined safety constraints. "Agent damage" refers to any harm to the agent, such as collisions or malfunctions. "Episode termination" occurs when the agent reaches a terminal state without successfully completing its assigned task.

The source environment is designed to mimic real-world scenarios but with reduced complexity and risk; i.e., a simulated environment where RL agent can explore the world and enter unsafe states with no real cost. Once the agent has navigated the source environment enough, we extract the safe state sequences it encountered during its exploration.

**2- Learn Safe Sequences:**
We use the collected safe state sequences to train a sequential anomaly detection model. This model aims to identify anomalous state sequences in the target environment. Each sequence of states is represented as a multivariate time-series which is a timestamped sequence of observations/states of size $T$:

$$\mathbf{X} = [\mathbf{s}_1, \mathbf{s}_2, \ldots, \mathbf{s}_T],$$

Where each state $\mathbf{s}_t$ is collected at a specific timestamp $t$, $\mathbf{s}_t \in \mathbb{R}^M$ where $M$ is the number of state variables in the environment state space and $T$ is the lookback timestamps, meaning that each sequence contains the past $T$ timesteps of state variables, this allows the model to capture temporal dependencies in the data and improve its ability to learn the safe sequences and detect anomalous states in the target environment.

We employ a transformer auto-encoder architecture, inspired by the work presented in [26] to reconstruct the safe sequences. The reconstruction loss is calculated as the mean absolute error (MAE):

$$\text{MAE} = \frac{1}{T} \sum_{t=1}^{T} |\hat{s}_t - s_t|$$

Where $\hat{s}_t$ is the reconstructed state at time $t$.
The anomaly score $\eta_t$ for a sequence at time $t$ is given by the reconstruction error:

$$\boldsymbol{\eta_t} = \sum_{t=1}^{T} \text{softmax}(-\frac{2}{\log(1 + \text{MAE})}) \cdot \text{MAE}$$

By using the transformer auto-encoder to learn a compact and informative representation of the safe sequences, we can effectively detect anomalies in the target environment and use this anomaly score as a risk measure to train a risk-averse policy in the target environment.

- **Phase 2: Train Risk-Averse Policy**

In this phase, we train a risk-averse policy in the target environment, which is a safety-critical environment (i.e., a real-world scenario). We use the anomaly scores from the sequence anomaly detection model as an estimated risk during online training in the target environment. We achieve this by adjusting the reward function to penalize the agent for visiting states that are detected as unsafe based on the anomaly detection output. Specifically, we add a risk penalty term to the standard reward function used in reinforcement learning. The risk penalty term increases with the anomaly score, discouraging the agent from entering potentially hazardous states.

The goal of the risk-averse RL actor-critic policy is to learn a policy that maximizes the expected cumulative reward while minimizing the risk of encountering anomalies. To achieve this, we modify the reward function to include the anomaly score, which encourages the agent to avoid states with high anomaly scores and discourages it from entering potentially hazardous states. Specifically, the reward function is defined as:

$$r_t = \begin{cases} r_t^{\text{orig}} & \text{if } \eta \leq \theta \\ r_t^{\text{orig}} - \beta \, \eta_t & \text{otherwise} \end{cases}$$

Where $r_t^{\text{orig}}$ is the original reward at time step $t$, $\eta_t$ is the anomaly score at time step $t$, $\theta$ is a threshold value for the anomaly score, and $\beta$ is a hyperparameter that controls the strength of the penalty term.

By combining sequential anomaly detection with reinforcement learning, our proposed approach enables the development of intelligent agents capable of operating safely in dynamic, uncertain environments. The next section presents experimental results demonstrating the effectiveness of our method.

## EXPERIMENTS

*A. Environments:*
To evaluate the performance of our proposed approach, we use three different benchmarking environments:

- **Safety Ant Run:** The environment, Safety Ant Run-v0, involves a quadrupedal agent with four legs. The task for the agent is to run through an avenue between two safety boundaries. Observations for the agent include sensory inputs regarding its position, velocity, and surrounding environment. Rewards are given for maintaining a running speed within the safety boundaries, while a cost signal is received when exceeding an agent-specific velocity threshold. [27].

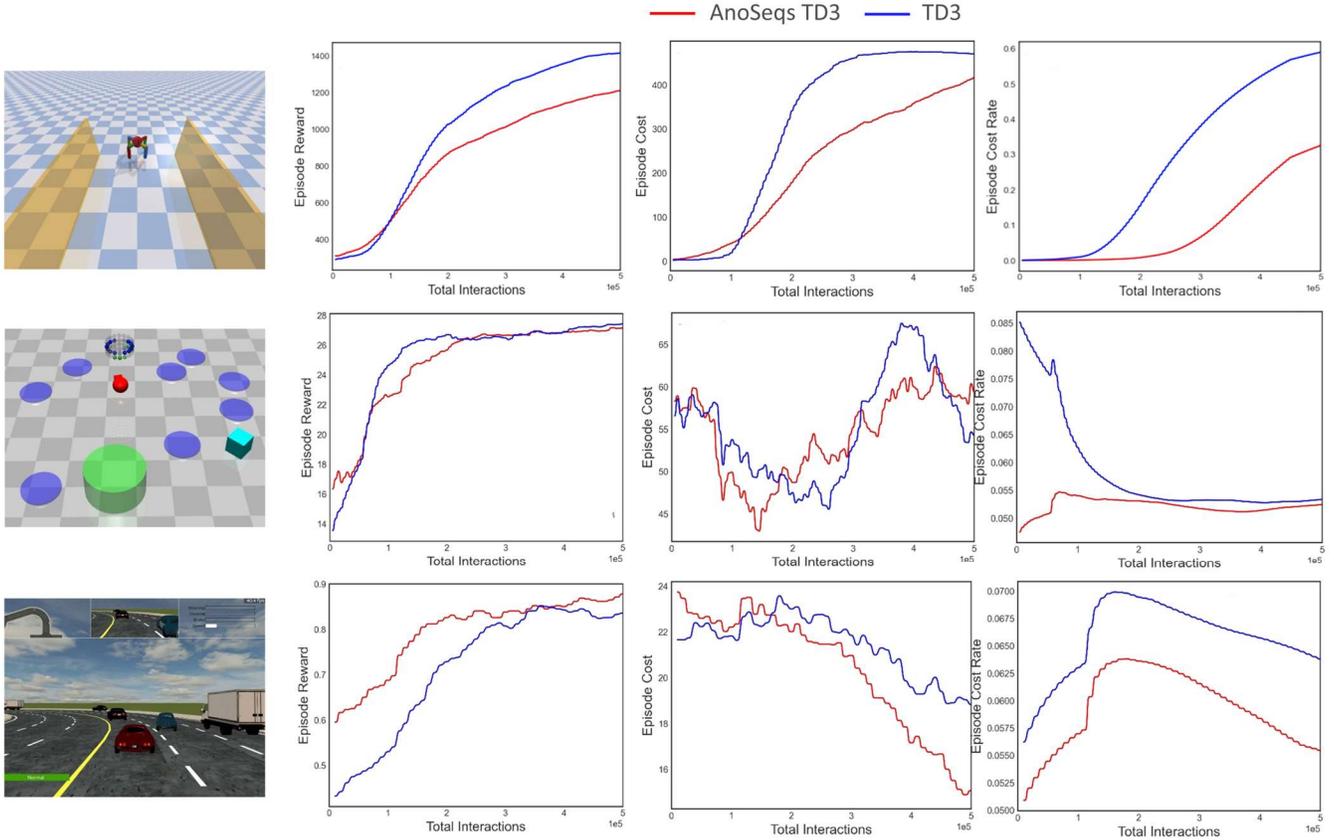

**FIGURE 2.** Comparison of learning curves for AnoSeqs and the baseline algorithm TD3 on the three benchmarking environments. Safety Ant Run (first row), Safety Gym Point Goal (middle row), and Safety Meta Drive (bottom row). The x-axis represents the number of interactions during the training process. The y-axes represent episodic return (left column), episodic cost rate (middle column), and total cost rate (right column), respectively. Red curves represent our modified algorithm, AnoSeqs, while blue curves represent the baseline algorithm, TD3. The results are averaged over multiple episodes and are captured at regular intervals during the training process.

- **Safety Gym Point Goal:** The environment, Safety Gym Point Goal-v1, is a mass point agent. The task for the agent is to navigate to the green goal while avoiding blue hazards. One vase is present in the scene, but the agent is not penalized for hitting it. Observations for the agent include data on its position, goal location, and positions of hazards. Rewards are provided for reaching the goal, and penalties are given for colliding with hazards [28].
- **Safety Meta Drive:** The environment, Safe Meta Drive, simulates an autonomous vehicle scenario. The task for the agent is to reach navigation landmarks as quickly as possible without colliding with other vehicles or going off-road. Observations for the agent include inputs regarding its speed, position, nearby vehicles, and road boundaries. Rewards are given for fast and safe navigation, while penalties are applied for collisions or going off-road [29].

These environments represent various safety-critical scenarios, such as driving within speed limits, avoiding pedestrians, and navigating through construction zones. We employ three metrics to measure the performance of our algorithm [30]:

### B. Metrics:
1. Episodic Return = sum of rewards in the test time. It indicates how well the agent finishes the original task.
2. Episodic Cost Rate = number of cost signals / length of the episode. It indicates how safe the agent is at the test time. A "cost signal" represents an event or state deemed undesirable or unsafe, such as a collision or safety violation.
3. Total Cost Rate = total number of cost signals / total number of training steps. It indicates how safe the agent is in the whole training process.

### HYPERPARAMETERS SETTINGS

In this study, we employed a sequential anomaly detection model based on a Transformer-based autoencoder to identify anomalous states in three different environments. The model was trained on safe state sequences collected using reinforcement learning agents, and we extracted safe sequences by checking for collisions, damage, or episode termination.

For the AnoSeqs algorithm, implemented using the twin-delayed deep deterministic policy gradient TD3 [31] policy within the SafeRL-Kit framework [30], we focused on two key hyperparameters: the anomaly threshold and the anomaly penalty. The anomaly threshold was set based on the 95th percentile of the mean absolute error for autoencoded safe sequences from the source environment.

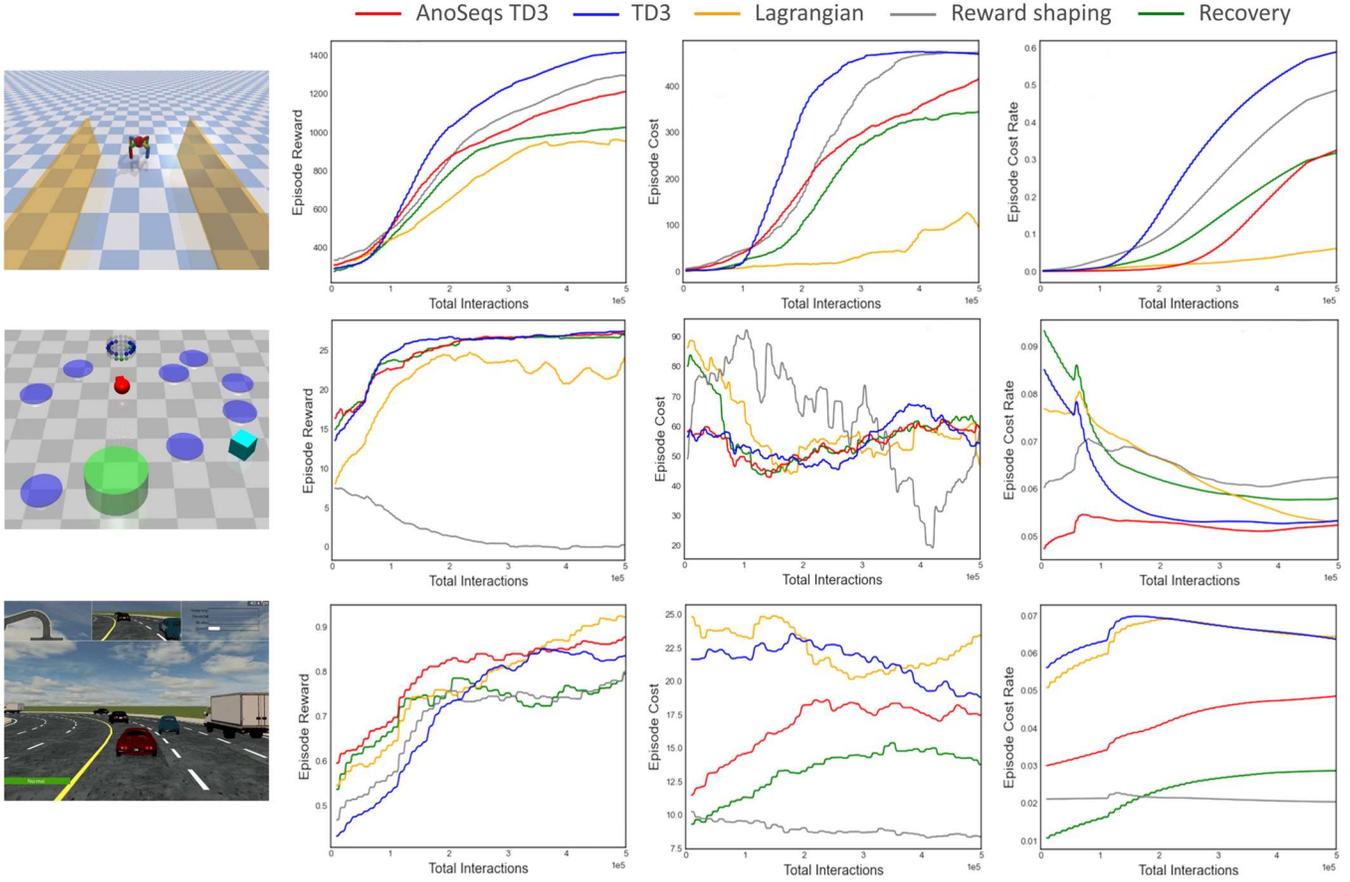

**FIGURE 3.** Comparison of learning curves for AnoSeqs and the three state-of-the-art algorithms: Lagrangian RL algorithm [1], Recovery RL algorithm [2], Reward Shaping algorithm [3, 4], and the baseline algorithm TD3 on the same three benchmarking environments. Safety Ant Run (first row), Safety Gym Point Goal (middle row), and Safety Meta Drive (bottom row). The x-axis represents the number of interactions during the training process. The y-axes represent episodic return (left column), episodic cost rate (middle column), and total cost rate (right column), respectively. Red curves represent our modified algorithm, AnoSeqs, while blue curves represent the baseline algorithm, TD3. The gray curves represent the Reward Shaping algorithm, the green curves represent the Recovery RL algorithm, and the yellow curves represent the Lagrangian RL algorithm.

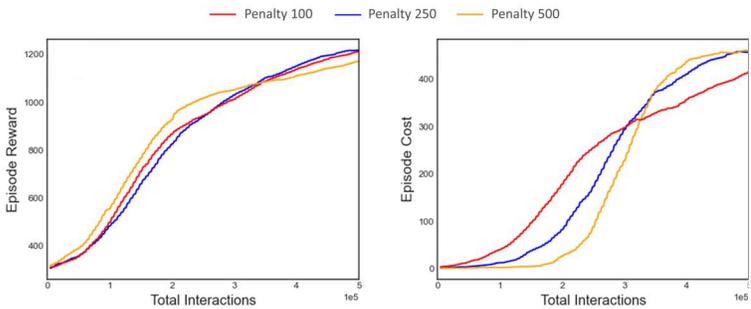

**FIGURE 4.** Sensitivity analysis on anomaly penalty

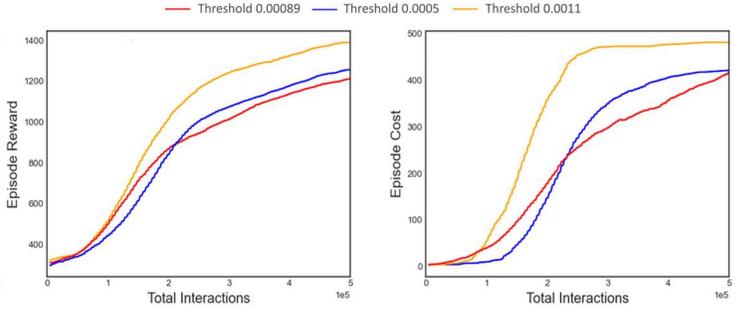

**FIGURE 5.** Sensitivity analysis on anomaly threshold

This threshold determined the sensitivity of the anomaly detection model, enabling the identification of anomalous state sequences while minimizing false positives. As for the anomaly penalty, we experimented with different penalties based on the anomaly score value and reward value. The goal was to strike a balance that discouraged the agent from entering anomalous states while encouraging exploration in safer areas of the state space. The specific values chosen for each environment are summarized in Table 1. These hyperparameter settings played a crucial role in shaping the behavior and performance of our algorithms during the experiments.

TABLE 1:
ANOSEQS HYPERPARAMETERS SETTINGS

| Environment | Anomaly threshold | Anomaly penalty |
|---|---|---|
| Safety Ant Run | 0.00089 | 100 |
| Safety Gym Point Goal | 0.00427 | 100 |
| Safety Meta Drive | 0.00247 | 10 |

# RESULTS

In this section, we will present the results of our proposed AnoSeqs method. We will compare its learning curves with baseline and state-of-the-art algorithms. Additionally, we will provide a summary of the algorithm performance after training and conduct a sensitivity analysis to examine the impact of different hyperparameter settings.

### A. Comparison with Baseline Algorithm TD3:

In this subsection, we present the results of our proposed method AnoSeqs and compare its performance with the baseline TD3 algorithm [31]. We evaluated our method on the three safety-critical benchmarking environments: Safety Ant Run, Safety Gym Point Goal, and Safety Meta Drive.

As shown in Figure 2, our proposed method, AnoSeqs, outperforms the baseline algorithm, TD3, in terms of reducing the cost rate during training, as well as the average episodic cost. The cost rate during training is consistently lower for AnoSeqs compared to TD3. In Safety Ant Run, the reduction in reward is expected as the algorithm becomes more cautious to avoid costly actions. In Safety Gym Point Goal, AnoSeqs achieves a lower cost rate during training and a slightly lower average episodic cost compared to TD3 but maintains a similar level of episodic reward, indicating that our algorithm is able to learn a safe policy without compromising the primary task. Moreover, in Safety Meta Drive, AnoSeqs not only reduces the cost rate during training but also increases the reward, due to the environment's reward design that includes negative signals for costly actions.

### B. Comparison with State-of-the-Art Algorithms:

As shown in Figure 3, the comparison results with other safety algorithms indicate that the AnoSeqs algorithm achieves a good balance between episodic return and cost. The state-of-the-art algorithms included in our comparison are the Lagrangian RL algorithm [1], Recovery RL algorithm [2], and Reward Shaping algorithm [3, 4], each ensuring safety at different stages of the reinforcement learning process. The Lagrangian RL algorithm incorporates safety during the optimization phase by enforcing constraints. The Recovery RL algorithm enhances safety at the output of the policy by introducing a recovery layer that mitigates unsafe actions. The Reward Shaping algorithms integrate safety into policy by penalizing costs.

In the Safety Ant Run environment, the AnoSeqs algorithm demonstrates a high episodic return (left column) compared to other algorithms, while maintaining a low episodic cost rate and total cost rate. Although the Lagrangian algorithm achieved the lowest cost rate, its return was low, indicating that this algorithm provides a high level of safety at the expense of performance due to its constraint satisfaction method. In the Safety Gym Point environment, the AnoSeqs algorithm achieves the highest episodic return with the lowest episodic and total cost rates. In the Safety Meta Drive environment, the algorithm also achieves high episodic returns with low episodic and total costs. The Recovery and Reward Shaping algorithms managed to reduce costs but at the expense of returns. These results highlight the ability of the AnoSeqs algorithm to sustainably improve performance across multiple environments.

### C. Summary Statistics of Algorithm Performance:

To provide a comprehensive evaluation of the AnoSeqs algorithm, we conducted 100 convergence tests and compared the results with the baseline TD3 algorithm, generating summary statistics post-training. This evaluation aims to demonstrate the robustness of the AnoSeqs algorithm, proving that it consistently maintains its performance after policy convergence.

TABLE 2:
SUMMARY STATISTICS OF ALGORITHM PERFORMANCE AFTER 100 CONVERGENCE TESTS

| Environment | Metric | TD3 Mean ± Std | AnoSeqs Mean ± Std |
|---|---|---|---|
| Safety Ant Run | Episode Cost | 470.95 ± 5.38 | 417.59 ± 5.77 |
| | Episode Rewards | 1430.98 ± 11.77 | 1189.09 ± 5.97 |
| Safety Gym Point Goal | Episode Cost | 53.11 ± 19.06 | 52.90 ± 17.09 |
| | Episode Rewards | 27.51 ± 0.48 | 27.22 ± 0.77 |
| Safety Meta Drive | Episode Cost | 19.67 ± 3.69 | 15.33 ± 3.41 |
| | Episode Rewards | 0.84 ± 0.07 | 0.80 ± 0.08 |

The results presented in Table 2 demonstrate that the performance metrics of the AnoSeqs algorithm post-convergence align closely with pre-convergence trends, specifically in minimizing costs while maintaining robust performance. Both the mean and standard deviation of episode costs and rewards across the evaluated environments remain consistent after 100 tests, highlighting the reliability and stability of the AnoSeqs algorithm. Furthermore, the low standard deviations across all metrics emphasize the robustness and consistency of AnoSeqs in delivering dependable performance.

### D. Sensitivity Analysis with Different Hyperparameter Settings:

We study the impacts of two critical hyperparameters in our model: the anomaly penalty ($\beta$) and the anomaly threshold ($\theta$) on the Safety Ant Run environment. The anomaly penalty $\beta$ controls the strength of the penalty term, while $\theta$ is a threshold value for the anomaly score. Figure 4 presents the impact of different penalty values on the Episode Reward and Episode Cost. The results show that even with higher penalties, which ensure more safety by minimizing risky actions, the learning curves maintain similar shapes, with lower costs and similar rewards, demonstrating robustness across different values. Through multiple experiments, we found that a penalty of 100 achieves the best balance between performance and safety.

For θ, the plots in Figure 5 demonstrate the sensitivity of the model to various threshold values. We tested different anomaly thresholds (0.00089, 0.0005, and 0.001) using different methods of calculating the anomaly threshold, such as the 95th percentile of the mean absolute error, 3 standard deviations from the mean, and 2 standard deviations from the mean. Lower thresholds increase the model's sensitivity to anomalies. As the threshold rises, the learning curves begin to resemble those of the baseline TD3 algorithm without anomaly modifications, indicating reduced sensitivity to anomalies. This suggests that with higher thresholds, the model may fail to detect anomalies, reverting to baseline performance. This analysis highlights the necessity of fine-tuning θ and β to balance the model's sensitivity to anomalies and the associated penalty, optimizing overall performance while maintaining safety.

Overall, the results demonstrate that our proposed method AnoSeqs is effective in enhancing the safety of reinforcement learning agents in various safety-critical environments and our algorithm is able to balance exploration and exploitation effectively while maintaining safety.

Our main contribution is the development of a novel method for learning to detect anomalous state sequences as the measure of unsafety, which enables any RL algorithm to become more cautious and safer. By incorporating anomalous state sequences into the training process, our algorithm can learn safer policies and reduce the likelihood of encountering undesirable events even without recovery, constraints, or safety projection techniques. Therefore, our method can be applied to safe RL algorithms in the literature to further enhance their safety.

## CONCLUSION

We proposed a novel approach to safe reinforcement learning called Safe Reinforcement Learning with Anomalous State Sequences (AnoSeqs). Our approach leverages sequential anomaly detection to enhance the safety of RL agents in dynamic and uncertain environments. By integrating anomalous state sequences into the training process by learning safe state sequences from a source environment, our algorithm is able to learn safer policies and reduce the likelihood of encountering unsafe actions. The experimental results demonstrated the effectiveness of our proposed method in various safety-critical environments.

Our work highlights the potential of sequential anomaly detection in safe reinforcement learning. By adopting this technique, RL agents can better handle unexpected events and learn safer policies in complex and dynamic environments. Future research directions may explore further integration of anomaly detection methods into different stages of the RL learning process, towards more robust and reliable AI systems.